\documentclass[letterpaper]{article}
\usepackage{aaai23}
\usepackage{times}  % DO NOT CHANGE THIS
\usepackage{helvet}  % DO NOT CHANGE THIS
\usepackage{courier}  % DO NOT CHANGE THIS
\usepackage[hyphens]{url}  % DO NOT CHANGE THIS
\usepackage{graphicx} % DO NOT CHANGE THIS
\usepackage{natbib}  % DO NOT CHANGE THIS AND DO NOT ADD ANY OPTIONS TO IT
\usepackage{caption} % DO NOT CHANGE THIS AND DO NOT ADD ANY OPTIONS TO IT
\usepackage{algorithm}
\usepackage{dcolumn}
\usepackage{algorithmic}
\usepackage{amsmath}
\usepackage{amssymb}
\usepackage{subfigure}
\usepackage{booktabs}
\usepackage{newfloat}
\usepackage{listings}
\newcommand{\tabincell}[2]{\begin{tabular}{@{}#1@{}}#2\end{tabular}} %放在导言区
\DeclareCaptionStyle{ruled}{labelfont=normalfont,labelsep=colon,strut=off} % DO NOT CHANGE THIS
\floatstyle{ruled}
\newfloat{listing}{tb}{lst}{}
\floatname{listing}{Listing}
\nocopyright
\title{BYHE: A Simple Framework for Boosting End-to-end Video-based Heart Rate Measurement Network}
\author{
    Weiyu Sun\textsuperscript{\rm 1}, Xinyu Zhang\textsuperscript{\rm 2}, Ying Chen\textsuperscript{\rm 2},\\
    Chunyu Ji\textsuperscript{\rm 3}, Yun Ge\textsuperscript{\rm 3}, Xiaolin Huang\textsuperscript{\rm 3}
}
\affiliations{
    \textsuperscript{\rm 1} Nanjing University\\
}

\usepackage{bibentry}
\begin{document}

\maketitle

\begin{abstract}
    Heart rate measuring based on remote photoplethysmography (rPPG) plays an important role in health caring, which estimates heart rate from facial video in a non-contact, less-constrained way. End-to-end neural network is a main branch of rPPG-based heart rate estimation methods, whose trait is recovering rPPG signal containing sufficient heart rate message from original facial video directly. However, there exists some easily neglected problems on relevant datasets which thwarting the efficient training of end-to-end methods, such as uncertain temporal delay and indefinite envelope shape of label waves. Although many novel and powerful networks are proposed, hitherto there are no systematic research digging into these problems. In this paper, from perspective of common intrinsic rhythm periodical self-similarity results from cardiac activities, we propose a comprehensive methodology, \textbf{B}oost \textbf{Y}our \textbf{H}eartbeat \textbf{E}stimation (BYHE), including new label representations, corresponding network adjustments and loss functions. BYHE can be easily grafted on current end-to-end network and boost its training efficiency. By applying our methodology, we can save tremendous time without conducting laborious handworks, such as label wave alignment which is necessary for previous end-to-end methods, and meanwhile enhance the utilization on datasets. According to our experiments, BYHE can leverage classical end-to-end network to reach competitive performance against those state-of-the-art methods on mostly used datasets. Such improvement indicates selecting perspicuous and efficient label representation is also a promising direction towards better remote physiological signal measurement.
\end{abstract}

\section{Introduction}
\begin{figure}
    \centering
    \setlength{\abovecaptionskip}{0.cm}
    \includegraphics[height=4.1cm]{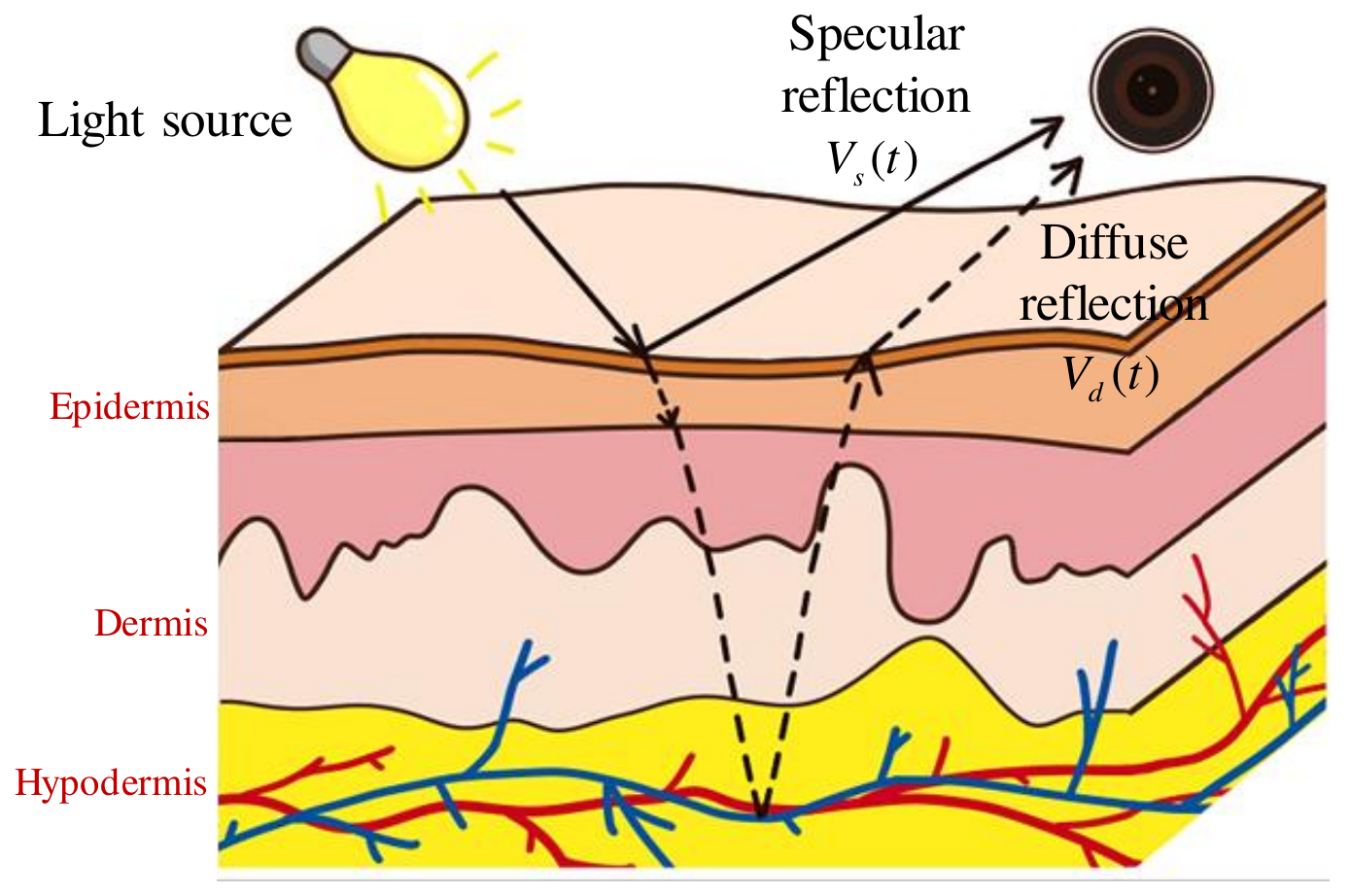}
    \caption{Vessels activity causes diffuse reflection's temporal change, it can be captured using single camera (i.e. webcam). By analyzing subtle color change (rPPG signals) $V_d(t)$, we can get heart rate information, because vessel activities result from heart activities. The picture refers from \cite{POS}.}
    \label{reflex_model}
\end{figure}

Non-contact video based-heart rate estimation established its theoretical foundation since last century, whose principle can be briefly described as Fig \ref{reflex_model}. Compared with traditional heart rate estimation methods such as electrocardiography (ECG) and Photoplethysmography (PPG), it's more convenient and safer for its unnecessity of physical contact. As realized as the extension of PPG, non-contact video based heart rate estimation is named as remote PPG (i.e. rPPG) method.

Early rPPG methods \cite{GREEN, CHROM, POS} didn't rely on deep learning tools. Based on prior experience concluded by researchers, we can craft a pipeline based on algorithms to realize remote heart rate estimation. By selecting region of interest (ROI, usually facial skin areas) per frame in facial videos and analyzing its color changes along time, we can generate a temporal sequence called rPPG signal. From rPPG signals we can easily extract heart rate information, just like what we do on ECG signals or Blood Volume Pulse (BVP) signals.

\begin{figure*}[t]
    \centering
    \setlength{\abovecaptionskip}{0.15cm}
    \includegraphics[width=13cm]{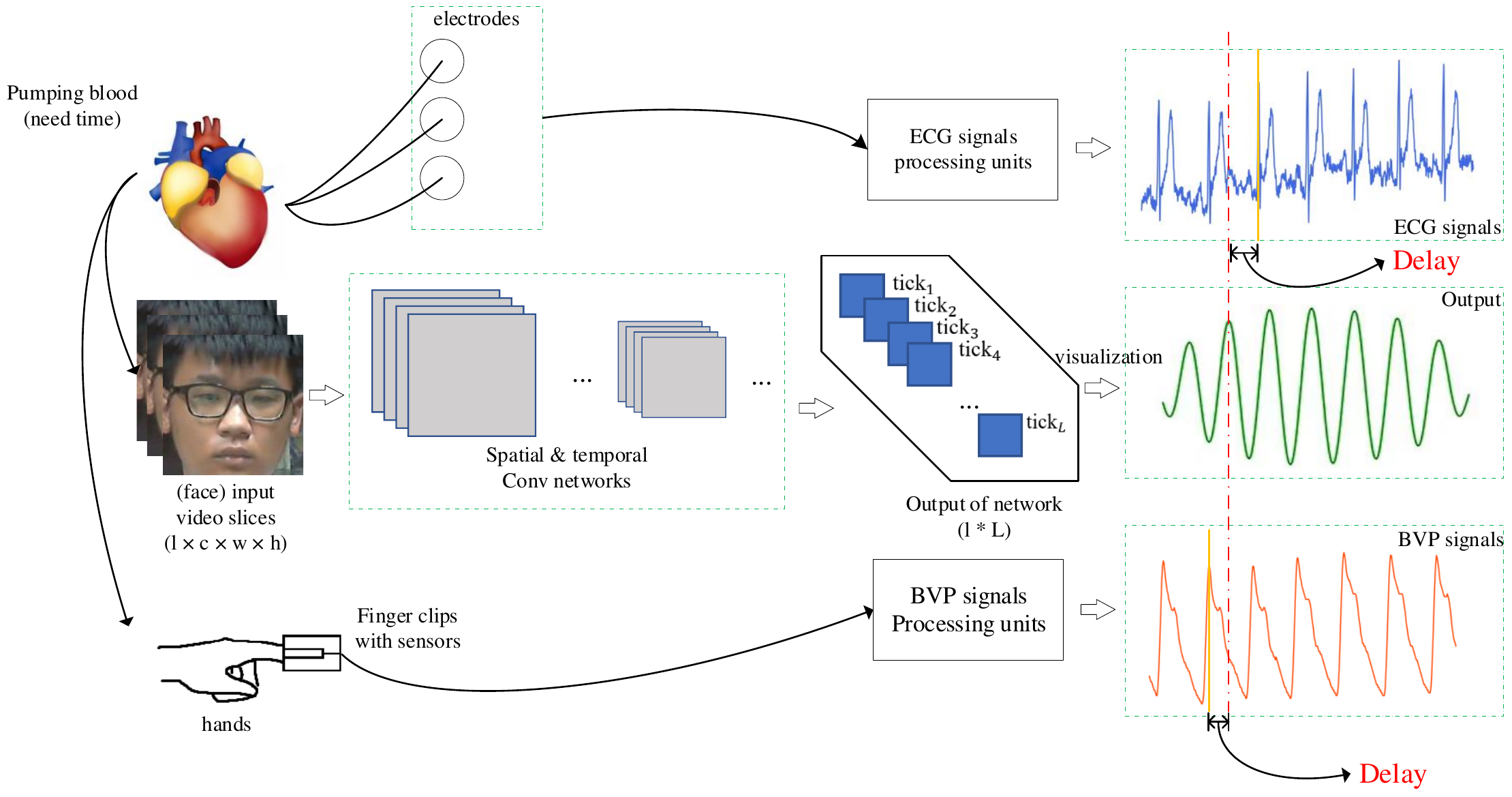}
    \caption{The specific process of different heart rate extraction method. ECG signals are collected through electrodes stick on body, they reflect cardiac activity. BVP signals are collected through finger clip sensors, they reflect contraction of finger vessels. Similarly, facial rPPG signals contain information on facial blood activity. Because of time consumption of blood flowing and unknown process time from signal processing units of BVP/ECG collecting devices, finally BVP/ECG signals and facial rPPG signals have uncertain temporal delay.}
    \label{Phase_delay phenomenon}
\end{figure*}

But prior concluded experience are not always perfectly suitable for every video. With deep learning taking off, researchers started using neural networks to assist rPPG signal extraction. These networks can be divided into two branches, end-to-end network and non end-to-end network. End-to-end network \cite{STVEN, Deepphys, Physformer} takes in facial videos and output predicted rPPG signals, and non end-to-end \cite{RhythmNet, CVD, Dual-GAN} output heart rate value directly. However, predicting heart rate straightly from raw facial videos is difficult for currently available network pipelines, so non end-to-end networks instead take in feature maps generated from facial videos. The generation process involves algorithm based methods, such as CHROM \cite{CHROM}. Each branch has its pros and cons, and their respective state-of-the-art methods share similar performance.

Meanwhile, relevant datasets are published. \textbf{However, almost all datasets' designers aren't so considerate for both two network branches mentioned above, especially end-to-end methods.} Main problem ties with ground truth labels of these datasets: they are usually ECG signals \cite{VIPL-HR} or BVP signals \cite{HCI}. From perspective of frequency, they are acceptable, ECG and BVP signals do contain sufficient heart rate information. But when considering time domain, ECG and BVP signals \textbf{are not temporally aligned with} facial rPPG signals, there exists uncertain delay between them. This phenomenon results from time consumption of physiological signal transmission (e.g. blood flowing) and device processing consumption, as shown in Fig \ref{Phase_delay phenomenon}. Besides, real rPPG signals' envelope shapes are indefinite: it is nearly impossible to observe QRS complex (ECG signal's feature, can be checked in Fig \ref{Phase_delay phenomenon}) on facial area through webcam. In addition, label waves' amplitudes rely on sampling systems, thus they are independent with amplitude of rPPG signals, so we have to instead focus on frequency $\&$ phase information of these label waves.

For end-to-end method whose output is predicted rPPG signal, we always hope its output is aligned with ground truth waves per frame, whereby we can easily design loss function (e.g. Negative Pearson correlation \cite{STVEN}) for it. Existence of uncertain delay prompts us to rethink the design of network, otherwise network can't converge (see Table \ref{不对齐不收敛}) under training. Nevertheless, problem on uncertain delay is not unsolvable, for example, we can use estimated rPPG signals from algorithm based method or previous trained end-to-end networks to serve as temporal standard for calibration. But applied methods are not reliable on all videos: they might generate wrong label waves which hinder efficient network training. Besides, alignment calibration process is \textbf{quite laborious}, when performing transfer learning and fine-tuning on new datasets, such calibration must be conducted again, which is struggling.

For non-end-to-end method, its training dosen't require all information from datasets: label waves are compressed into heart rate as output labels, facial videos are compressed into feature maps to let network easily learn on frequency features, while artificially discarding these information limits network performance. Non end-to-end performance relies on the quality of input feature maps, which are generated using classical algorithms \cite{CHROM}. According to experiments, these algorithms are not robust (see Table \ref{VIPL-HR结果} and \ref{MAHNOB-HCI结果}): they perform poor on noisy facial videos. To solve this problem, researchers make effort to improve the quality of generated input feature maps, such as splitting facial frames into small patches \cite{RhythmNet} or compressing different ROI combinations into feature maps \cite{CVD}. These methods provide more informative and detailed feature maps for subsequent heart rate extraction.

In this paper, we propose a simple but efficient training methodology, \textbf{B}oost \textbf{Y}our \textbf{H}eartbeat \textbf{E}stimation (BYHE). BYHE can be easily grafted on end-to-end pipelines, it liberates network trainers from annoying temporal alignment. According to experiments, BYHE is also conducive to higher performance of grafted end-to-end network. Compared to non end-to-end method, our method exploit information directly from input facial videos (the trait of end-to-end method), which is more potential. Noteworthy, BYHE is not to compete with previous methods, in contrast, it provides a heuristic tool. For example, our output matrix $\hat{R}$ can serve as inputs of non end-to-end methods, which is generated using network feature extractors, different from those extracted based on algorithms.

\section{Related works}
% \subsection{Algorithm-based method}
% Mature theory foundations \cite{Euler-Enlarge, POS} admit us to build rPPG extraction algorithms without deep learning tools. Many details in algorithms are based on experience assumptions, such as pixels of green channels contains more rPPG information according to the absorption spectrum of skin; analyzing pixel changes of background to offset the influence of ambient light change; using fixed ROI such as forehead and cheek areas instead of whole face because these areas are relatively more informative. Although diverse, almost all algorithm based method can be described as following steps: (1) Gain facial videos sequence and extract skin areas \cite{skin_seg_rgb} of selected ROI per frames. (2) calculating statistical pixel intensity values (usually calculating average values of all pixel in ROI) per channel per frame, thereby gain temporal intensity sequences of different color channels. (3) Merge all intensity sequences into one sequences using predefined methods \cite{PCA, ICA, GREEN} (e.g. choosing green channel sequences only). (4) filter this sequence and conclude its frequency, which is the final heart rate's frequency.

% The assumptions of these algorithms are sufficient enough to predict precise heart rate, under common condition. While accidents always takes place, such as hair covers forehead. And deep learning methods mentioned below can enhance the robustness of rPPG extraction.

\subsection{Non End-to-end method}
Non end-to-end methods develop both on input feature maps and network structures. Input feature maps are quite diverse, such as the short-time-Fourier-transform spectrums of rPPG signal \cite{short-Fourier2017}, synthesized signals made using CHROM \cite{CHROM} from 25 patches of raw videos \cite{RhythmNet}, different combinations of ROI places on facial area \cite{CVD}. All these settings are to improve the input quality for better network processing. As to the network backbone, it ranges from ResNet18 (altered into regression structure) \cite{zhangsenle, SynRhythm, RhythmNet}, encoder-decoder structure \cite{CVD} to generative adversarial network (GAN) structure \cite{Dual-GAN}. Better feature representation along with better backbones result into higher performance of non End-to-end method.
% Non End-to-end methods \cite{RhythmNet, CVD, zhangsenle, Dual-GAN} don't output predicted rPPG signals, instead, they output single heart rate values. By doing this, non end-to-end methods shun the uncertain delay problem of relevant datasets. On the other hand, these networks don't use facial videos as input, their inputs are well-designed feature maps made from facial videos. The process of feature maps generation involves classical methods mentioned above, so inevitably some noise will be contained into generated feature maps. Feature maps with poor quality make networks unable to extract usefully information to infer the true heart rate. Because of the limitation of classical methods, we couldn't generate good qualitied feature maps on every video, especially those high-noised videos.

% Some skills are applied to enhance the information quality of generated feature maps. For example, RhythmNet's \cite{RhythmNet} inputs contain additional color channels (HSV). CVD's inputs consist of different combinations of facial areas to maximize the information. These skills could be of some help to some extent, but once some processes of classical methods are used, such as give skin areas in ROI the same weights when calculating averaging pixel values, the problem could not be uprooted.

\subsection{End-to-end method}
The evolution on end-to-end method focuses on the network structure. Early end-to-end methods used convolutional nerual network (CNN) as network backbone, from 2D CNN version \cite{Deepphys} to 3D CNN version \cite{CAN, STVEN, PhysNet}. These CNN structures are lightweight, generally comprise of a few CNN layers. Early end-to-end methods were less competent than non end-to-end methods, some researchers even applied neural architecture search strategy \cite{NAS} to seek for the most optimized structure. Recently, with more nouveau network structures proposed, the end-to-end backbone gets upgraded concomitantly. Such as meta learning stragegy \cite{Meta-rppg} and transformer structure \cite{Physformer}, these new methods rift the performance of this branch comparable with that of non end-to-end method.

But according to our investigation, few relevant papers explain how they preprocess label waves, such as methods on aligning label waves. Although some methods apply tricks such as using loss function on frequency domain \cite{Physformer}, but directly comparing predicted wave and label wave is still necessary on these methods. Hitherto, dealing with dataset-related problems such as temporal misalignment and indefinite wave envelope are still less researched. Next, we will show researches into this field is worthy and promising.
% End-to-end method\cite{Deepphys, STVEN, Physformer}, on the other hand, uses facial videos as input and outputs predicted rPPG signals. Compared to the pipeline of non end-to-end methods, end-to-end methods in essence use networks to extract facial rPPG signals directly, without the necessity to make feature maps according to certain inductive bias. After proper training, networks rPPG information extraction can be more flexible and robust. However, as pointed above, relevant datasets don't befit the network structure well: the uncertain delay is quite a big deal. Because out goal is to make the network's output approach real rPPG signal, so it is quite hard for trainers to gain the true temporal-aligned ground-truth rPPG signals! But we still have methods. For example, we could use classical methods or previously trained end-to-end methods to generate ``label rPPG signals''. This is plausible but not all videos in relevant datasets could generate reliable label signals, the quality of these label signals are quite pool when input videos are highly noisy. Meanwhile, re-calibrate these label signals is laborious. When conducting transfer learning and finetuning on new datasets, such struggling but inevitable process still need to be done.

% Under such circumstance, end-to-end method still shows its competence on heart rate estimation: its up-to-date method hitherto has similar performance with non end-to-end's newest method.

\section{Approach}
Facial color changes (rPPG signals), BVP signals and ECG signals have common intrinsic connection, they all result from cardiac activities. Although they generally present different waveform patterns (such as QRS complex of ECG signals and quadratic harmonic peak of BVP signals), they share the same rhythm and frequency which reflect the information of heart rate. We exploit the inherent law of their rhythms and then design methods to extract these rhythm features. In this chapter, we introduce our proposed method -- BYHE, including preprocessing method on ground truth waves (BVP signals and ECG signals), adjusted network structure, designs of corresponding loss functions and the calculation of heart rate. Despite simple, BYHE can help network trainers train network efficiently and harvest higher performance.

We utilize self-attention mechanism which focuses on the internal periodicity on rhythms of rPPG signals and ground truth waves instead of their situation in specific time stamps. Thereby BYHE excludes the effect of uncertain delay between ground truth waves and real facial color changes, and circumvents indefinite envelope shape problem of label waves.

\subsection{Label representation}
\begin{figure}
    \centering
    \setlength{\abovecaptionskip}{0.1cm}
    \includegraphics[height=4.3cm]{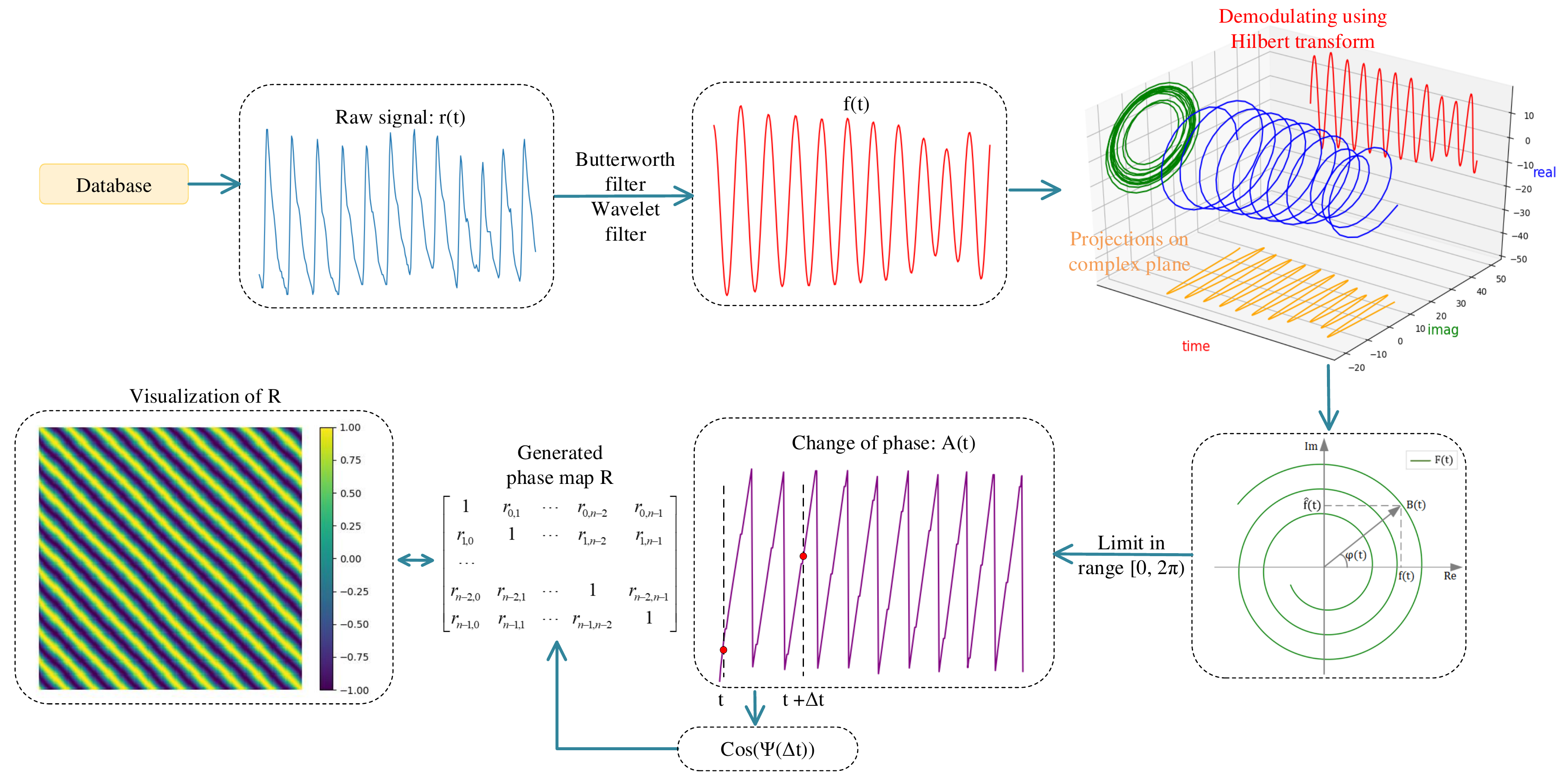}
    \caption{
    Preprocessing on datasets (BVP signal in this figure). When demodulating $f(t)$, we utilize Hilbert transform to generate $\hat{f}(t)$ (yellow curve in complex plane), and we focus on the helical line (green circles) projected by $F(t)$ on the complex plane. The helical line in figure is transformed in order to explicitly show the instantaneous phase $\varphi(t)$. Purple line are A(t) integrated by $\varphi(t)$ and it is wrapped between $[0, 2\pi)$, every index in whose x-axis reflects a special time stamp recorded in ticks (let all waves in figure have totally n ticks). Then all pairs of time stamps will be calculated $\cos{({\Psi}({\Delta }t))}$ and inserted into the phase map R according to their indexes. Final visualization of phase map shows the periodicity and rhythm properties of ground truth wave.
    }
    \label{preprocessing}
\end{figure}

Generally, ground truth waves can be divided into BVP signals (such as VIPL-HR datasets \cite{VIPL-HR}) and ECG signals (such as MAHNOB-HCI datasets \cite{HCI}). In this session, we display proposed label representation generation method (from BVP signals) in Fig \ref{preprocessing}. \textbf{Methods for ECG signals could be found in Apppendix.}

Human's heart rate frequency is generally between 0.7 Hz and 4 Hz (conservative estimate). Therefore, first we apply Butterworth filter within [0.7, 4] Hz and continuous wavelet transform (i.e. CWT) filter \cite{cwt_filter} on ground trouth wave $r(t)$ to exclude unrelevent noise. \textbf{The implementation of CWT filter (including parameters) can be found in Appendix.1} After filtering, the frequency of filtered wave $f(t)$ becomes monolithic, thus $f(t)$ could be described as equation (\ref{filteredft}):
\begin{equation}
    f(t)=B(t)\cos{(2{\pi}f_{s}t + \varphi)}
    \label{filteredft}
\end{equation}
where $\varphi$ is uncertain delay between rPPG signals and ground truth waves (ECG or BVP signals). $f_{s}$ is the main frequency of heart rate in $r(t)$, and $B(t)$ is the envelope of the signal depends on the amplitude of $r(t)$, $B(t)$'s frequency is much lower than $f_{s}$. The form of $f(t)$ in (1) resembles modulated signals in signal transmission, and we only cast attention on $2{\pi}f_{s}t + \varphi$ in $f(t)$ because the envelope $B(t)$ is independent of facial condition. Therefore, we apply Hilbert transform \cite{Hilbert} to demodulate $f(t)$ and extract $2{\pi}f_{s}t + \varphi$ and $B(t)$ separately. The demodulation with Hilbert transform could be briefly demonstrated as (\ref{Hilbert}):
\begin{equation}
    % \begin{aligned}
    F(t)=f(t)+i\hat{f}(t)=\overline{f}e^{i{\varphi}(t)}
    % \end{aligned}
    \label{Hilbert}
\end{equation}
$$where \ \overline{f}={\sqrt{f(t)^{2} + \hat{f}(t)^{2}}}, \  {\varphi}(t)=arctan(\hat{f}(t)/f(t))$$
In (\ref{Hilbert}), $\hat{f}(t)$ is the result Hilbert transform of $f(t)$. Considering the frequency of envelope $B(t)$ is much lower than $f_{s}$, the $\varphi(t)$ of the analytical signal $F(t)$ can be approximated described as (\ref{analytical}):
\begin{equation}
    {\varphi}(t)\approx arctan(\frac{B(t)sin(2{\pi}f_{s}t + \varphi)}{B(t)cos(2{\pi}f_{s}t + \varphi)})=2{\pi}f_{s}t + \varphi
    \label{analytical}
\end{equation}
Thereby we gain the instantaneous phase change $\varphi(t)$ of $f(t)$. After integrating $\varphi(t)$ along time, we thereby get temporal phase change $A(t)$. By far, the uncertain delay $\varphi$ still exists. To uproot the influence of $\varphi$, we cast attention on the difference ${\bigtriangleup}t$ between frames by calculating the cosine value of $\Psi_{{\Delta}t}$, which could be described as (\ref{draw-varphi}):
\begin{equation}
    \begin{aligned}
        \Psi_{{\Delta}t} & =A(t+{\Delta}t)-A(t) \\&=((2{\pi}f_{s}(t+{\Delta}t)-2k\pi + \varphi)-(2{\pi}f_{s}t + \varphi))\\&=2{\pi}f_{s}{\Delta}t
    \end{aligned}
    \label{draw-varphi}
\end{equation}
where $2k\pi$ is the wrapped period in $A(t)$. After subtraction, the common component $\varphi$ in $A(t+{\Delta}t)$ and $A(t)$ is neutralized, hence we exclude the uncertain delay $\varphi$. We calculate $cos(A(t_{1})-A(t_{2}))$ on each combination of time stamps $t_{1},t_{2} \in \{ 0,1,2,\dots, n-1\} $, where n is the length of $r(t)$. Thereby we gain the target label matrix ${R_{n\times n}}$, in which every element ${r_{ij}} = \cos{(\Psi_{i-j})}$. The visualization of ${R_{n\times n}}$ is is shown in Fig \ref{preprocessing}.

Compared with single heart rate used in non end-to-end network training, ${R_{n\times n}}$ reserves more temporal information. Combining with adjusted network structure and corresponding loss functions mentioned below, we can perform efficient end-to-end learning not affected by uncertain delay, which can't be realized by using single heart rate as label.

\subsection{Network structure}
Now let's adjust traditional end-to-end network, make it output $\hat{R}$ sharing similar property with $R$. Specifically, we reserve wrapped end-to-end network pipeline except the last layer (i.e. global average pool). Then we use sliding windows with fixed strides (we use stride 1) to collect feature map slices ${s_0, s_1, \cdots, s_{N-1}}$ along time dimension, where N is number of total feature map slices. After flattening and going through feature projection network (i.e. linear layers), each feature map slices $s_i$ are turned into feature vectors $v_i$, which contains temporal rPPG context information of $s_i$. Similarly with the generation of $R$, we calculate cosine similarity on all combinations $(v_{i}, v_{j})$, then input into output matrix $\hat{R}$, where element $\hat{R}_{i,j}$ represents the cosine similarity value between $v_{i}$ and $v_{j}$. \textbf{The adjusted structure is shown in Fig \ref{adjusted-network}. More detailed information (including whole training flowchart) can be checked in Appendix.2} Though adjusted structure is simple, it's sufficient for us to conduct our methodology. The periodicity of facial color changes causes the repeatability of local rhythm patterns, which is recorded in generated vector sets ${v_{0}, v_{1}, \cdots, v_{N-1}}$. Therefore, by calculating cosine similarity among generated vectors, we could observe this periodicity through each row and column in $\hat{R}$ which swings with the increasing of distance between two vectors in time scale, shown in Fig \ref{adjusted-network}.

% \begin{figure}
%     \centering
%     \includegraphics[height=6cm]{adjusted network.pdf}
%     \caption{Adjusted network structures, output $\hat{R}$ and used to calculate loss function with $R$ using equation (\ref{loss-func}).}
%     \label{adjusted-network}
% \end{figure}

\begin{figure*}[t]
    \centering
    \setlength{\abovecaptionskip}{0.15cm}
    \includegraphics[height=6.5cm]{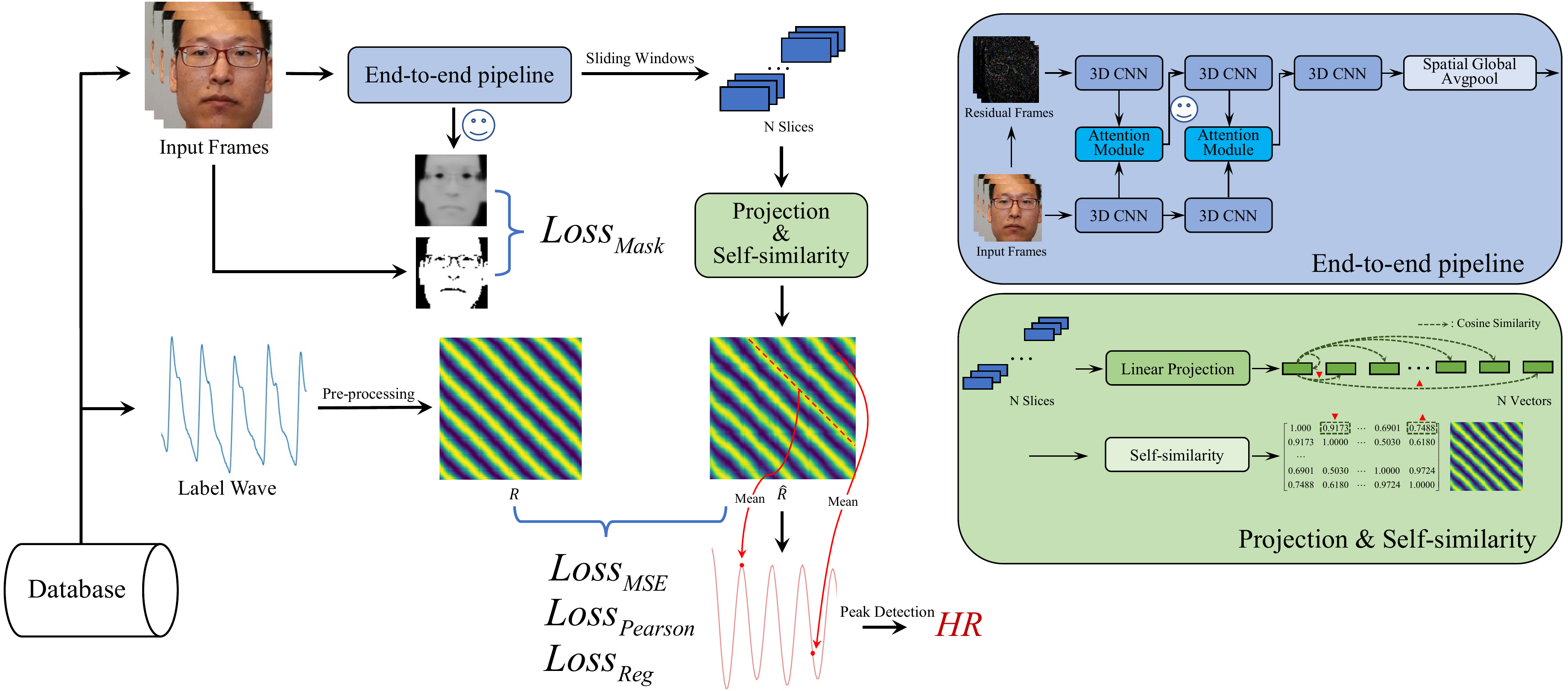}
    \caption{Adjusted network structures (green box), output $\hat{R}$ will be used to calculate loss function with $R$ using equation (\ref{loss-func}) for heart rate extraction.}
    \label{adjusted-network}
\end{figure*}

\subsection{Training process}
Now we gain two matrix $\hat{R}$ and $R$, we generate them using two individual systems and they are subtly different on physical meaning. But they can both represent heart rate's periodicity and rhythm information, and they share the same value range (the range of cosine function). Here, we make an inductive assumption that output $\hat{R}_{i,j}$ is close to $R_{i,j}$ on each element. We measure similarity between them and design loss functions. Our experiments ascertain the reasonability of our assumption. Total loss functions can be described as follows:
\begin{equation}
    Loss = \alpha * Loss_{MSE} + \beta * Loss_{Pearson} + \gamma * reg
    \label{loss-func}
\end{equation}
where $Loss_{MSE}$ is Mean Square Error between $\hat{R}_{i,j}$ and $R_{i,j}$, and $Loss_{Pearson}$ is negative Pearson loss between them (mean negative Pearson loss between each row of $\hat{R}$ and $R$). \textbf{$reg$} represents regularization norm we set, which will be introduced in next session. It can help make the periodicity feature of $\hat{R}$ more stable. According to ablation study, optimized configuration is: $\alpha=1$, $\beta=0.8$ and $\gamma=0.1$.

\subsection{Heart rate estimation}
Output matrix $\hat{R}_{i,j}$ contains sufficient heart rate information, it resembles feature maps of non end-to-end methods, as shown in Fig \ref{adjusted-network}. According to observation, $\hat{R}_{i,j}$ contains very single frequency (viewed as heart rate), we can use simple method to extract frequency from it. Specifically, We collect elements $\hat{r}_{i,j}$ as groups $g_a$ where $abs(j-i)=a$ and $a\in[0, N-1]$, $N$ is the size of $\hat{R}$. Then calculate average value $m_a$ on each group $g_a$. Thus we gain 1-D sequence ${m_0, m_1,..., m_{N-1}}$. Apparently, high qualitied $\hat{R}$ have its elements in $g_i$ more uniformed, which indicates slices between the same time interval have similar relationship. So, we proposed regularization norm $reg$ as equation (\ref{reg}). The less the standard deviation of elements in each $g_i$ is, more uniformed the $\hat{R}$ is.
\begin{equation}
    reg = \frac{\sum_{i=0}^{N-1}{std(g_i)}}{N}
    \label{reg}
\end{equation}
After [0.7, 4] Hz Butterworth bandpass filter and CWT filter \cite{cwt_filter} (same configuration with data processing) on ${m_0, m_1,..., m_{N-1}}$, we calculate average peak intervals as heart rate's periodicity, as shown in Fig \ref{adjusted-network}. Of course, we can also use non end-to-end methods, such as RhythmNet \cite{RhythmNet}, let $\hat{R}$ as input and output heart rate directly. According to experiments, simple peak detection method is sufficient for generic heart rate estimation.

\section{Experiments}
\subsection{Experiment Setup}
We use altered CAN \cite{CAN} as grafted end-to-end network backbone. CAN evolved from Deepphys \cite{Deepphys} (3D CNN version of Deepphys), and Deepphys was claimed as first related end-to-end method. CAN was proposed in 2020, so it's quite a suitable backbone to exhibit the effect of BYHE: it's classical enough and not so up-to-date. We select VIPL-HR \footnote{\url{https://vipl.ict.ac.cn/zygx/sjk/201811/t20181129_32716.html}} \cite{VIPL-HR} and MAHNOB-HCI \footnote{\url{https://mahnob-db.eu/hci-tagging/}} \cite{HCI} as involved datasets, and perform 5-fold and 3-fold \textbf{subjected-exclusive} cross validation on them respectively, following the configuration of previous works \cite{Physformer, RhythmNet}. Both datasets are widely used and have many recorded performance of previous methods, and they are public available (if asked). We use ``mmod face detector'' \cite{mmod} from python package ``dlib'' to detect face in videos. When training, we input 70-frames facial sequence. When testing, we input 300-frames facial sequence. All frames in facial sequence are cut according to the first detected face detection box. We train BYHE on Tesla v100 with batchsize 6. We use adam optimizer \cite{adam} with dynamic learning rate \cite{COSINE_SGD}. We use random horizontal/vertical flipping and temporally up/down-sampling for data augmentation, following PhysFormer \cite{Physformer}, along with random crop. For metrics, we select mean absolute error (MAE), standard deviation (std) and root mean square error (RMSE) to quantify precision between label and predicted heart rates. Lower these metrics, better the performance of methods. We use peak detection method to extract heart rate from label waves when conducting validations. All metrics are measured on top of beats per minute (i.e. bpm). \textbf{More details can be found in Appendix.3}

\begin{table}
    \centering
    \setlength{\abovecaptionskip}{0.15cm}
    \resizebox{1.05                                                             \columnwidth}{!}{
        \begin{tabular}{l|ccc}
            \hline
            \multicolumn{1}{c|}{Method}                     & \tabincell{c}{std                                       \\(bpm)}   & \tabincell{c}{MAE\\(bpm)}   & \tabincell{c}{RMSE\\(bpm)}            \\
            \hline
            $\blacktriangle$Tulyakov2016\cite{Tulyakov2016} & 18.0              & 15.9             & 21.0             \\
            $\blacktriangle$POS \cite{POS}                  & 15.3              & 11.5             & 17.2             \\
            $\blacktriangle$CHROM \cite{CHROM}              & 15.1              & 11.4             & 16.9             \\
            \hline
            $\blacklozenge$RhythmNet\cite{RhythmNet}        & 8.11              & 5.30             & 8.14             \\
            $\blacklozenge$ST-Attention\cite{ST-Attention}  & 7.99              & 5.40             & 7.99             \\
            $\blacklozenge$CVD\cite{CVD}                    & 7.92              & 5.02             & 7.97             \\
            $\blacklozenge$Dual-GAN\cite{Dual-GAN}          & \underline{7.63}  & \underline{4.93} & \textbf{7.68}    \\
            \hline
            $\star$I3D\cite{I3D}                            & 15.9              & 12.0             & 15.9             \\
            $\star$PhysNet\cite{PhysNet}                    & 14.9              & 10.8             & 14.8             \\
            $\star$DeepPhys\cite{Deepphys}                  & 13.6              & 11.0             & 13.8             \\
            $\star$AutoHR\cite{NAS}                         & 8.48              & 5.68             & 8.68             \\
            $\star$PhysFormer\cite{Physformer}              & 7.74              & 4.97             & 7.79             \\
            $\star$\textbf{BYHE (Ours)}                     & \textbf{7.59}     & \textbf{4.85}    & \underline{7.73} \\
            \hline
        \end{tabular}
    }
    \caption{Performance on VIPL-HR dataset, the results of previous methods share the same protocol as claimed in \cite{Physformer, RhythmNet}. The data we refered comes from the latest paper \cite{Physformer}. Alike, symbols $\blacktriangle$, $\blacklozenge$, $\star$ denote traditional, non-end-to-end learning based and end-to-end learning based methods, respectively. Best results
        are marked in bold and second best in underline.}
    \label{VIPL-HR结果}
\end{table}

\begin{table}
    \centering
    \setlength{\abovecaptionskip}{0.15cm}
    \resizebox{1.05\columnwidth}{!}{
        \begin{tabular}{l|cccc}
            \hline
            \multicolumn{1}{c|}{Method}                      & \tabincell{c}{std                                       \\(bpm)}   & \tabincell{c}{MAE\\(bpm)}   & \tabincell{c}{RMSE\\(bpm)}            \\
            \hline
            % $\blacktriangle$Poh2011\cite{Poh2010}            & 13.5  & -     & 13.6  \\
            $\blacktriangle$CHROM\cite{CHROM}                & -                 & 13.49            & 22.36            \\ %CHROM
            $\blacktriangle$Li2014\cite{Li2014}              & 6.88              & -                & 7.62             \\
            $\blacktriangle$Tulyakov2016\cite{Tulyakov2016}  & 5.81              & 4.96             & 6.23             \\
            \hline
            $\blacklozenge$SynRhythm\cite{RhythmNet-prelude} & 10.88             & -                & 11.08            \\
            $\blacklozenge$RhythmNet\cite{RhythmNet}         & 3.99              & -                & 3.99             \\
            \hline
            $\star$HR-CNN\cite{HR-CNN}                       & -                 & 7.25             & 9.24             \\
            $\star$rPPGNet\cite{STVEN}                       & 7.82              & 5.51             & 7.82             \\
            $\star$DeepPhys\cite{Deepphys}                   & -                 & 4.57             & -                \\
            $\star$AutoHR\cite{NAS}                          & 4.73              & 3.78             & 5.10             \\
            $\star$Meta-rPPG\cite{Meta-rppg}                 & 4.9               & \textbf{3.01}    & \textbf{3.68}    \\
            $\star$PhysFormer\cite{Physformer}               & \textbf{3.87}     & \underline{3.25} & \underline{3.97} \\
            $\star$\textbf{BYHE (Ours)}                      & \underline{3.97}  & 3.54             & 4.11             \\
            \hline
        \end{tabular}
    }
    \caption{Performance on MAHNOB-HCI dataset.}
    \label{MAHNOB-HCI结果}
\end{table}

\subsection{Experiment Result}
\begin{figure}
    \centering
    \setlength{\abovecaptionskip}{0.cm}
    \subfigure[]{\includegraphics[width=4.16cm]{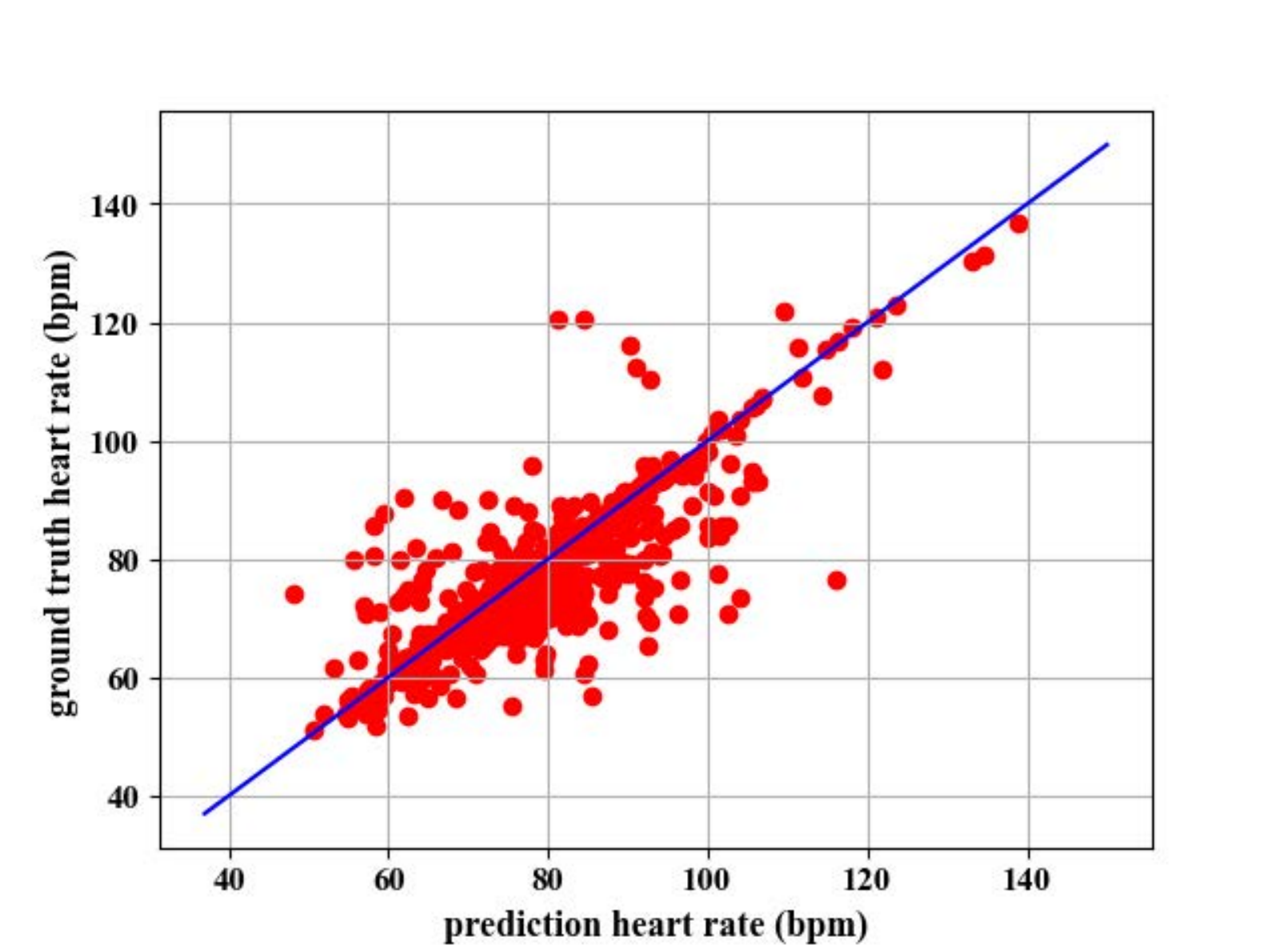}}
    \subfigure[]{\includegraphics[width=4.16cm]{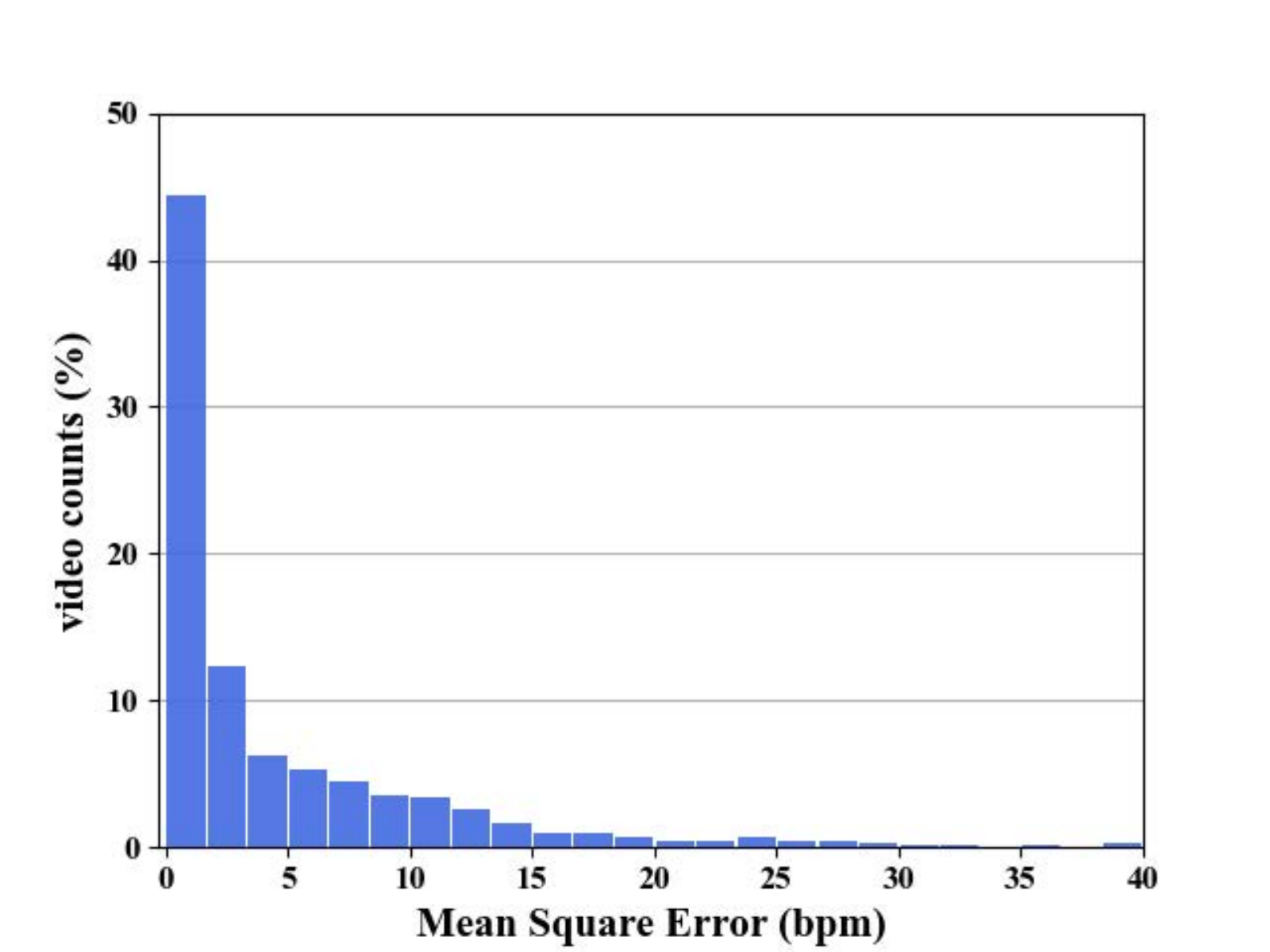}}
    % \subfigure[CHROM]{\includegraphics[width=4cm]{Tri_scatter_CHROM.eps}}
    \caption{Distribution of prediction results (bpm) and target results (a) and error (MAE) (b)} %图片标题
    \label{scatter-plot}
    % \label{fig:1}  %图片交叉引用时的标签
\end{figure}

Performance of BYHE along with previous methods are shown in Table \ref{VIPL-HR结果} and Table \ref{MAHNOB-HCI结果}. According to results, BYHE can reach comparable performance with other state-of-the-art methods, especially on VIPL-HR. We must point out, compared with other end-to-end methods involved in Table \ref{VIPL-HR结果} and Table \ref{MAHNOB-HCI结果}, \textbf{BYHE gains extra advantage on not performing struggling label wave temporal alignment.} To explicitly exhibit the error (MAE) distribution of our method, we plot distribution of results in source2 from VIPL-HR, as shown in Fig \ref{scatter-plot}.
% Performance of our method along with previous methods are shown in Table \ref{VIPL-HR结果} and Table \ref{MAHNOB-HCI结果}. We use altered CAN \cite{CAN} as wrapped end-to-end network backbone. Main reason for choosing this backbone is that its output are calculated from residual frames (i.e. next frame subtract current frame), so it would be natural zero-mean no matter the intensity of ambient light. This is our ideal data distribution for our self-cosine similarity calculation in adjusted network structure, then without wasting any bias adjustment in networks, the value of cosine similarity can be spread within [-1, 1] if rPPG signals are successfully extracted.

% According to results, BYHE gain comparable performance with other state-of-the-art methods. Here we must point out, compared with other end-to-end methods involved in Table \ref{VIPL-HR结果} and Table \ref{MAHNOB-HCI结果}, \textbf{our methods quite save lots of time for temporal alignment of label waves.} To explicitly exhibit the error distribution of our method, we plot distribution of results in source2 from VIPL-HR, as shown in Fig \ref{scatter-plot}.

\subsection{Ablation Study \footnote{All ablation experiments involved on VIPL-HR and HCI are conducted on folder2.}}
Some readers may argue that BYHE takes advantage of selected end-to-end pipeline. It can't be denied but besides enhancing training convenience of grafted end-to-end network, BYHE can also improve its performance. As shown in Table \ref{不对齐不收敛} and Fig \ref{曲线下降图}, BYHE rifts inner end-to-end pipeline to higher performance on two datasets. Noteworthy, as mentioned above, not all videos can be converted into reliable standard waves for temporal alignment. So when conducting experiments without BHYE on end-to-end methods, \textbf{we only select available video slices after temporal alignment} as train set, which is reasonable in practice. Meanwhile, Table \ref{不对齐不收敛} also indicates that end-to-end network can't reach convergence if we don't train it on aligned datasets. Therefore, when training end-to-end network or perform transfer learning on new datasets, considering BYHE is a good choice for convenience and performance.

\begin{table}
    \centering
    \setlength{\abovecaptionskip}{0.15cm}
    \resizebox{1.\columnwidth}{!}{
        % \resizebox{\linewidth}{!}{
        \begin{tabular}{c|c|c|ccc}
            \hline
                    & \tabincell{c}{BYHE Applied} & \tabincell{c}{Label Aligned} & \tabincell{c}{std                                 \\(bpm)}   & \tabincell{c}{MAE\\(bpm)}   & \tabincell{c}{RMSE\\(bpm)}            \\
            \hline
                    & $\times$                    & $\times$                     & -                 & -             & -             \\
            VIPL-HR & $\times$                    & \checkmark                   & 8.76              & 5.89          & 9.01          \\
                    & \checkmark                  & $\times$                     & \textbf{8.01}     & \textbf{4.82} & \textbf{8.08} \\
            \hline
            \hline
                    & $\times$                    & $\times$                     & -                 & -             & -             \\
            HCI     & $\times$                    & \checkmark                   & 5.67              & 4.35          & 5.81          \\
                    & \checkmark                  & $\times$                     & \textbf{4.31}     & \textbf{3.78} & \textbf{4.37} \\
            \hline
        \end{tabular}
    }
    \caption{End-to-end networks can't converge directly on VIPL-HR and HCI: label waves must be aligned. In contrast, our method can train directly and gain better performance. ``-" means failed.}
    \label{不对齐不收敛}
\end{table}

\begin{figure}
    \centering
    \setlength{\abovecaptionskip}{0.1cm}
    \includegraphics[height=3.5cm]{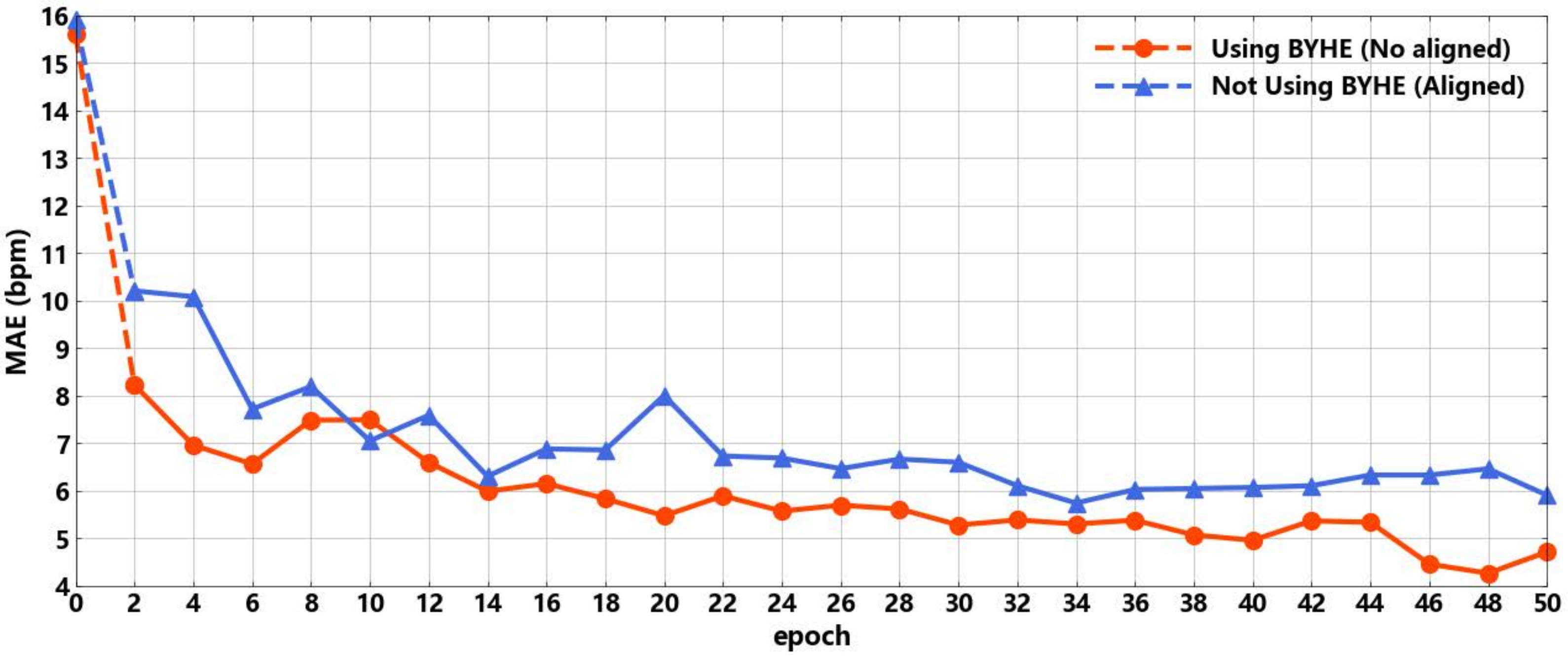}
    \caption{Performance of different training strategies on VIPL-HR.}
    \label{曲线下降图}
\end{figure}

In BYHE, sliding window length decides the observation length for each temporal slice $s_i$, thereby impacts the effect of mutual cosine similarity calculation and final performance of BYHE. We perform ablation experiments to seek the best observation length for similarity calculation. Our grafted end-to-end pipeline \cite{CAN} involves temporal convolution, so we additional stick temporal convolution length along with sliding window's length in ablation results. Results are shown in Table \ref{步长超参数对比}.

\begin{table}
    \centering
    \setlength{\abovecaptionskip}{0.15cm}
    \begin{tabular}{c|ccc}
        \hline
        \tabincell{c}{Sliding window                                \\ length} & std(bpm)       & MAE(bpm)      & RMSE(bpm)      \\
        %   & (bpm)          & (bpm)         & (bpm)          \\
        \hline
        1           & 8.78          & 5.43          & 8.81          \\
        6           & 8.12          & 4.98          & 8.23          \\
        \textbf{11} & \textbf{8.01} & \textbf{4.82} & \textbf{8.08} \\
        18          & 8.36          & 5.20          & 8.37          \\
        25          & 9.01          & 5.66          & 9.23          \\
        \hline
    \end{tabular}
    \caption{Performance on different sliding window lengths. Extra temporal convolution receptive field of inner end-to-end network is \textbf{15}. The best length ($11+15$ frames, $fps$ is 30) closes to the common period of cardiac activity (i.e. 70 bpm).}
    \label{步长超参数对比}
\end{table}

% \begin{table}
%     \centering
%     \setlength{\abovecaptionskip}{0.15cm}
%     \begin{tabular}{c|c|ccc}
%         \hline
%         % $L_{MSE\_map}$  & L & L & L & L & L & MAE(bpm) \\
%         Structure & Layers & \tabincell{c}{MAE \\(bpm)}   & \tabincell{c}{std\\(bpm)}   & \tabincell{c}{RMSE\\(bpm)}            \\
%         %    & (bpm) & (bpm) & (bpm)           \\
%         \hline
%         % & (bpm)  & std & RMSE  \\
%         \tabincell{c}{Single                   \\ Projection}                & $[88 \times 88]$                 & \textbf{4.82}              & \textbf{7.83}   & \textbf{7.88}   \\
%         \hline
%         \tabincell{c}{Single                   \\ Projection}  & $[88 \times 352]$                         & 5.17            & 8.27 & 8.26 \\
%         \hline
%         \tabincell{c}{Bottleneck               \\Projection} & \tabincell{c}{$[88 \times 1024]$                     \\$[1024 \times 1024]$\\$[1024 \times 88]$}    & 5.06  & 7.85 & 7.91\\
%         % -                & -        & -        & -         \\
%         \hline
%     \end{tabular}
%     \caption{
%         % $L_{map}$ is $L_{Pear\_map}\&L_{MSE\_map}$, $L_{wave}$ is $L_{Wave\_map}
%         % \&L_{Wave\_map}$, 
%         Performance on different projection structure.}
%     \label{backbone消融实验}
% \end{table}

\begin{table}
    \centering
    \setlength{\abovecaptionskip}{0.15cm}
    \begin{tabular}{c|c|ccc}
        \hline
        % $L_{MSE\_map}$  & L & L & L & L & L & MAE(bpm) \\
        Structure & Layers & \tabincell{c}{std \\(bpm)}   & \tabincell{c}{MAE\\(bpm)}   & \tabincell{c}{RMSE\\(bpm)}            \\
        %    & (bpm) & (bpm) & (bpm)           \\
        \hline
        % & (bpm)  & std & RMSE  \\
        \tabincell{c}{Single                   \\ Projection}                & $[88 \times 88]$                 & \textbf{8.01}              & \textbf{4.82}   & \textbf{8.08}   \\
        \hline
        \tabincell{c}{Single                   \\ Projection}  & $[88 \times 352]$                         & 8.27            & 5.17 & 8.26 \\
        \hline
        \tabincell{c}{Bottleneck               \\Projection} & \tabincell{c}{$[88 \times 1024]$        \\$[1024 \times 1024]$\\$[1024 \times 88]$}    & 8.15  & 5.06 & 8.17\\
        % -                & -        & -        & -         \\
        \hline
    \end{tabular}
    \caption{
        % $L_{map}$ is $L_{Pear\_map}\&L_{MSE\_map}$, $L_{wave}$ is $L_{Wave\_map}
        % \&L_{Wave\_map}$, 
        Performance on different projection structure.}
    \label{backbone消融实验}
\end{table}

\begin{table}
    \centering
    \setlength{\abovecaptionskip}{0.15cm}
    \resizebox{1.\columnwidth}{!}{
        \begin{tabular}{c|c|c|c|c|c}
            \hline
            % $L_{MSE\_map}$  & L & L & L & L & L & MAE(bpm) \\
            Case              & $L_{MSE}$  & $L_{Pearson}$ & $L_{mask}$ & $Reg$      & \tabincell{c}{MAE \\(bpm)}           \\
            \hline
            w/o MSE           & -          & \checkmark    & \checkmark & \checkmark & 6.35              \\
            w/o Pearson       & \checkmark & -             & \checkmark & \checkmark & 4.88              \\
            % MSE$\&$Pearson & \checkmark & \checkmark    & -               & -          & 5.23 \\
            \hline
            w/o mask          & \checkmark & \checkmark    & -          & \checkmark & 4.94              \\
            w/o reg           & \checkmark & \checkmark    & \checkmark & -          & 5.12              \\
            w/o reg $\&$ mask & \checkmark & \checkmark    & -          & -          & 5.14              \\
            default           & \checkmark & \checkmark    & \checkmark & \checkmark & \textbf{4.82}     \\
            \hline
        \end{tabular}
    }
    \caption{
        % $L_{map}$ is $L_{Pear\_map}\&L_{MSE\_map}$, $L_{wave}$ is $L_{Wave\_map}
        % \&L_{Wave\_map}$, 
        Ablation study on loss function items on VIPL-HR (folder 2), $L_{mask}$ is optional loss function binding with our selected end-to-end backbones, which could be checked in Fig \ref{adjusted-network}.}
    \label{Loss函数消融实验}
\end{table}

\begin{figure}
    \centering
    \setlength{\abovecaptionskip}{0.cm}
    \subfigure[]{\includegraphics[width=4.16cm]{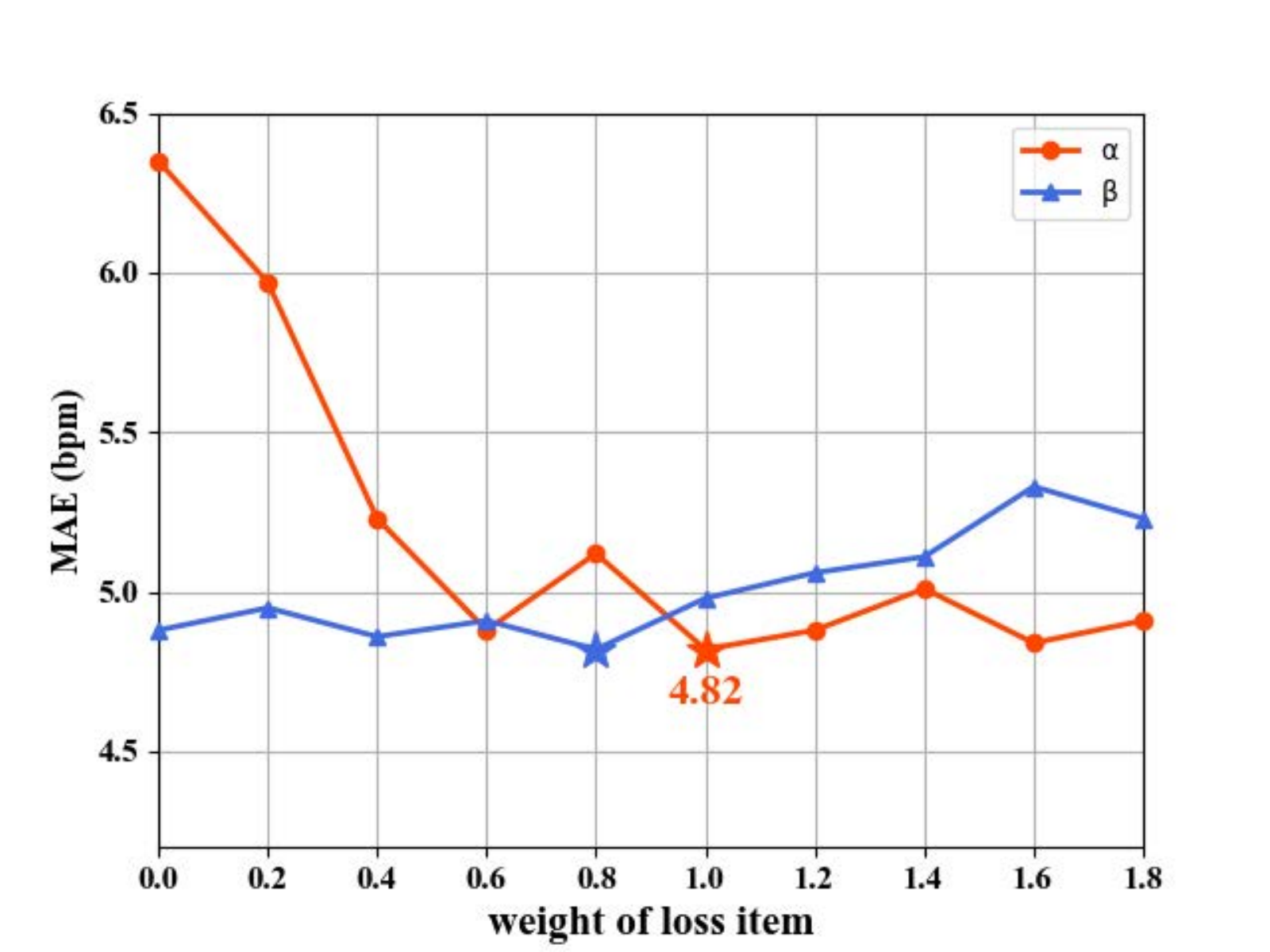}}
    \subfigure[]{\includegraphics[width=4.16cm]{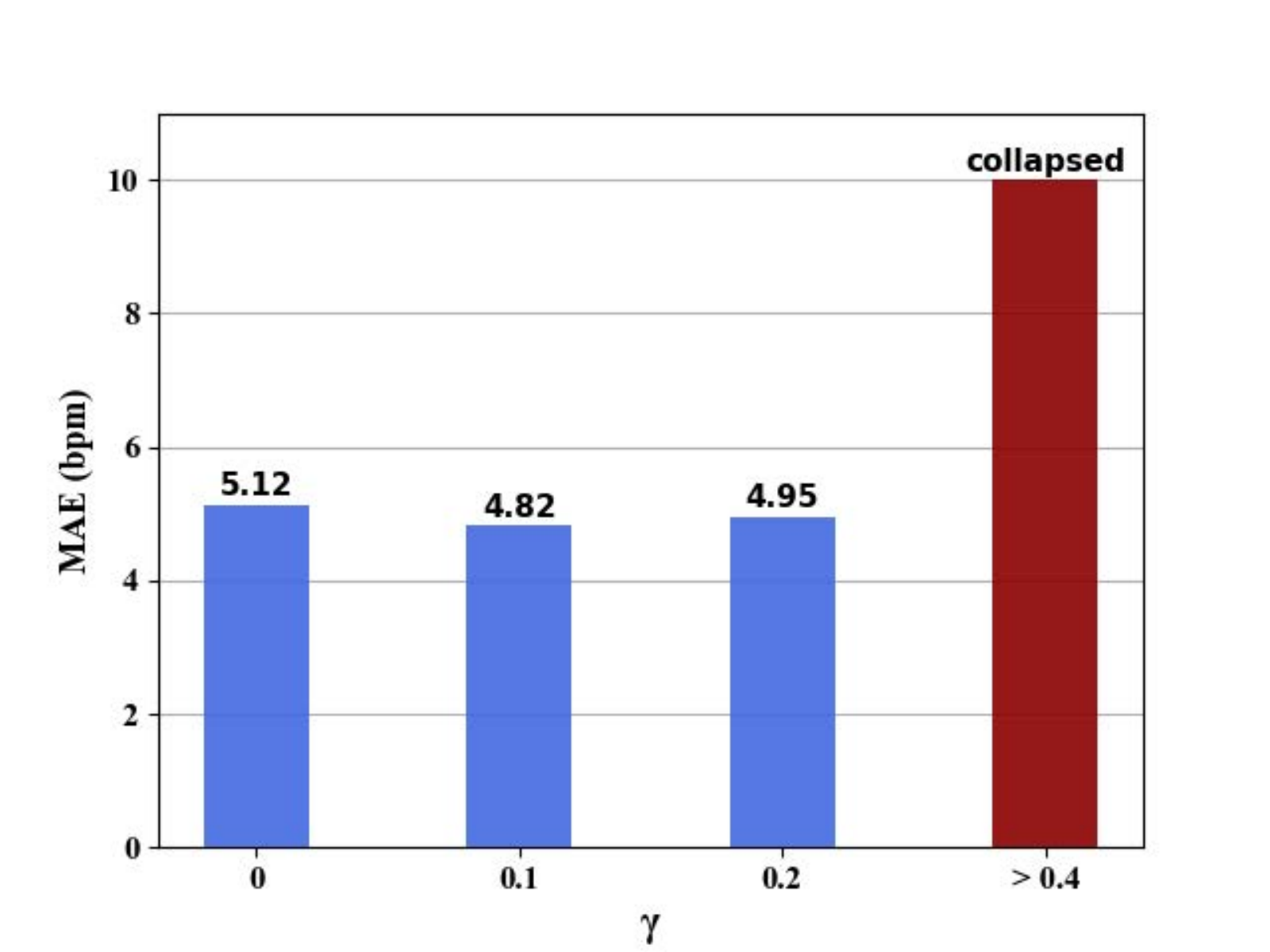}}
    % \subfigure[CHROM]{\includegraphics[width=4cm]{Tri_scatter_CHROM.eps}}
    \caption{Ablation study on weight of loss items (a) and regularization norm (b) in equation (\ref{loss-func}), ``collapse'' means all element in $\hat{R}$ becomes 1, where we can't extract heart rate from $\hat{R}$.} %图片标题
    \label{Loss权重系数消融}
    % \label{fig:1}  %图片交叉引用时的标签
\end{figure}

The pattern information extracted by end-to-end feature extractor is recorded in feature map slices ${s_0, s_1, \cdots, s_{N-1}}$. According to our experiment (Table \ref{backbone消融实验}), generally these rhythm patterns are clear enough: even a single linear projection layer can process these patterns into effective feature vectors ${v_0, v_1, \cdots, v_{N-1}}$ for subsequent cosine similarity calculation.

Our Loss function (equation (\ref{loss-func})) mainly consists of three items. $L_{MSE}$ focus on the similarity on each element between $R$ and $\hat{R}$, and $L_{Pearson}$ control the rhythm consistency between them. $Reg$ is set to make $\hat{R}$ more diagonal uniformed. We conduct experiments to search the best combination of these items. Noteworthy, different end-to-end pipelines have their own designed loss functions, they cause the diversity of equation (\ref{loss-func}), so we briefly exhibit the proposed common items on it. The ablation results are listed in Table \ref{Loss函数消融实验} \footnote{The weight of $L_{Mask}$ is 0.2. Available on all experiments involved $L_{Mask}$.} and Fig \ref{Loss权重系数消融}. According to results, though $L_{MSE}$ dominate for higher performance of BYHE, suitable $L_{Pearson}$ and $Reg$ can further enhance the effect. Giving huge weight for $Reg$ causes network collapse: it too much focuses on the uniformity of $\hat{R}$ so that network makes all elements in $\hat{R}$ into constant (i.e. 1). Results also prove BYHE's label and output representation catch the intrinsic commonplace between label waves and rPPG signals. By simply comparing the value between each element in $R$ and $\hat{R}$ (i.e. using $L_{MSE}$ only), we can just approach the most optimized effect of BYHE.

\subsection{Discussion}
\begin{figure}
    \centering
    \setlength{\abovecaptionskip}{0.cm}
    \subfigure[]{\includegraphics[width=4.16cm]{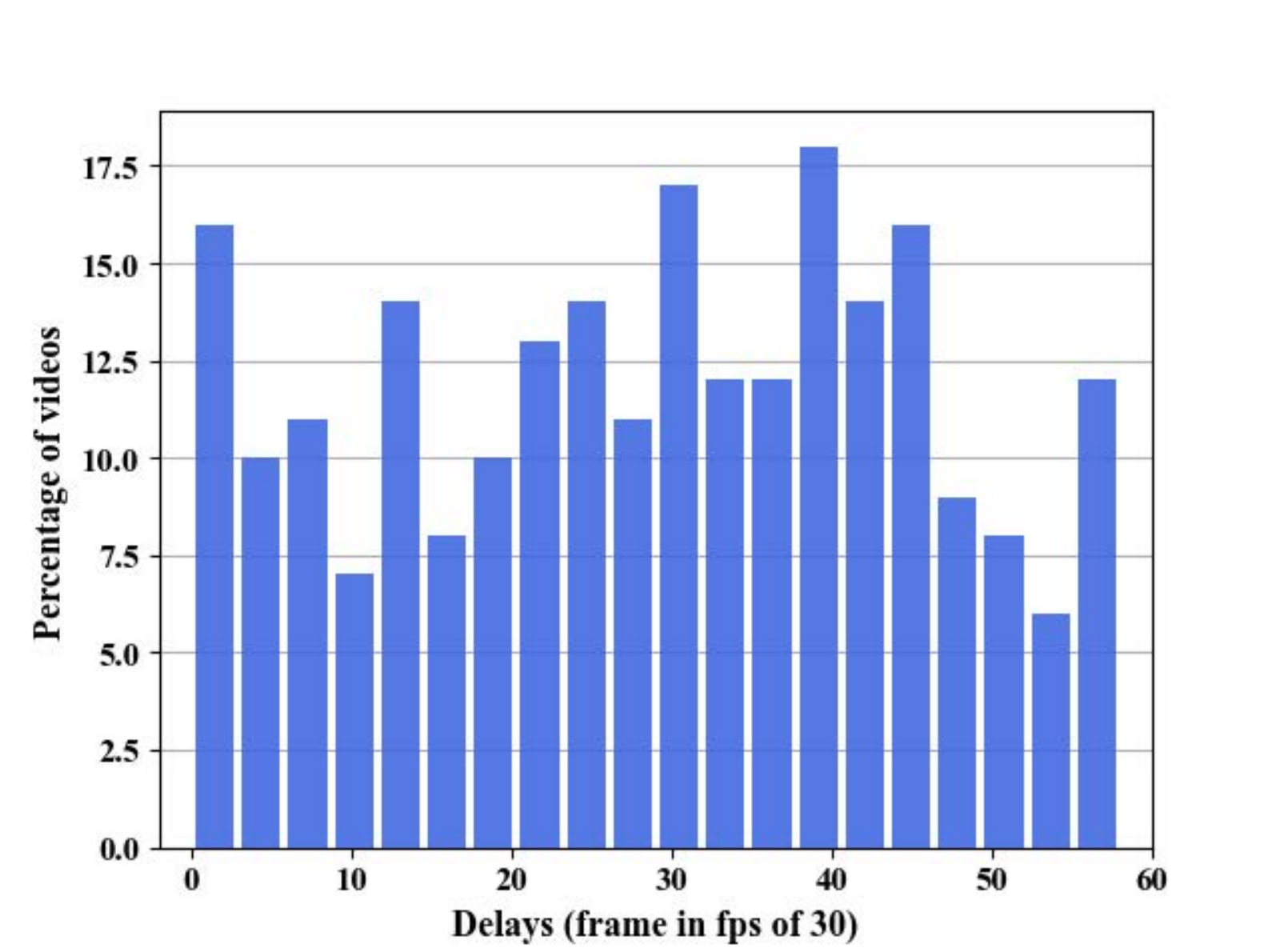}}
    \subfigure[]{\includegraphics[width=4.16cm]{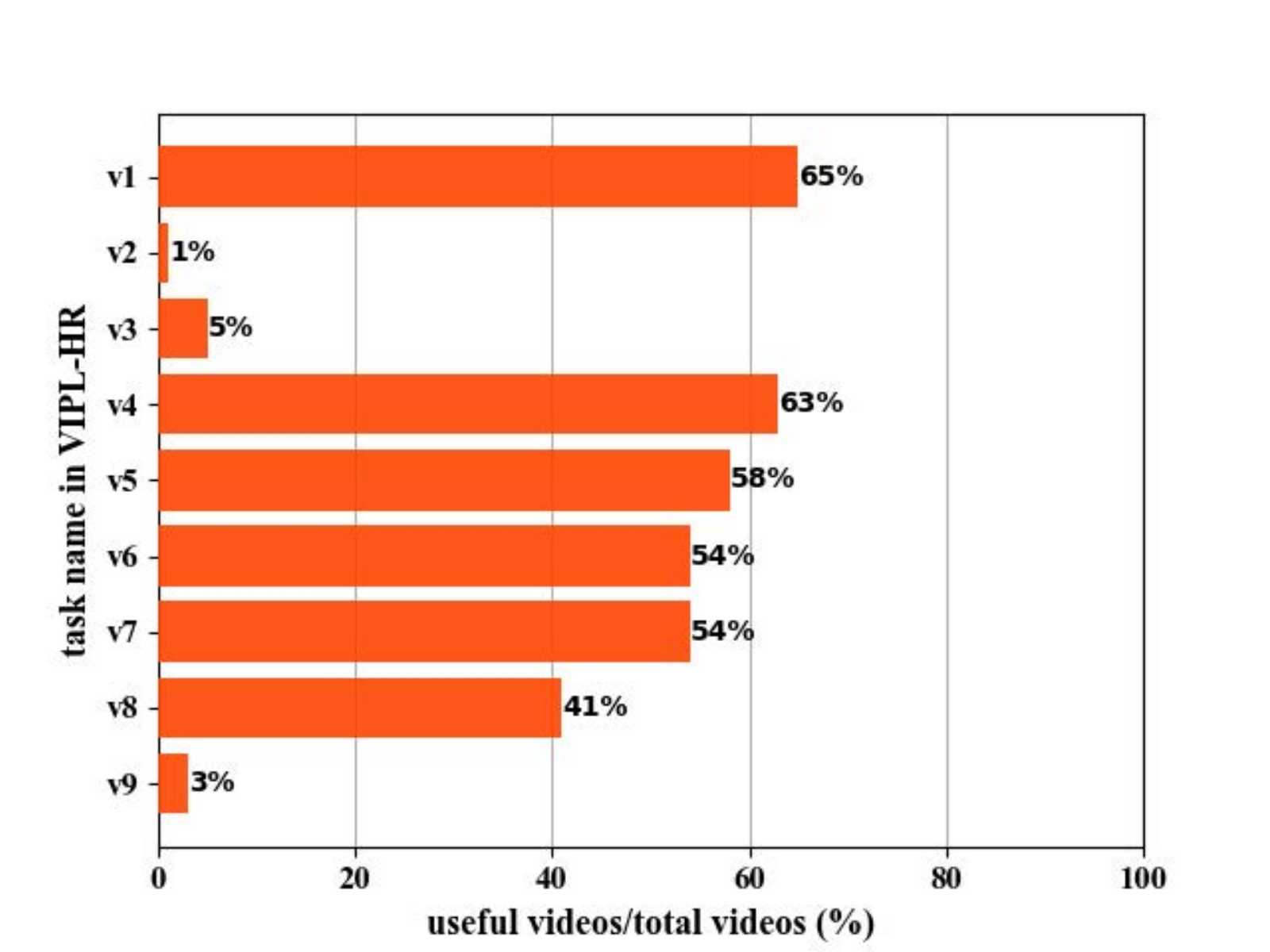}}
    \caption{(a) Delays' distribution on videos which can provide reliable standard waves using CHROM, samples come from participants 1 to 60 in VIPL-HR (source1 to 3). Delays are quantified using frames (30 frames per second). We can see only few ground truth waves are naturally temporal aligned with true rPPG signals in videos. (b) data utilization after alignment (v1 to v9, source 1 to source 3 in VIPL-HR).}
    \label{Distribution-of-source2}
\end{figure}

Researchers have dug deeply into the performance of end-to-end network pipelines, from 3D CNN to transformer-like structure. But input and output of these structures are almost unchanged, and problems on datasets mentioned above are not clearly discussed. Some paper simply claim that ground truth waves haven been aligned before training \cite{PhysNet}. In effect, performing alignment is far laborious than expected, and it means we have to compromise on the utilization on datasets. Fig \ref{Distribution-of-source2} shows available video slices we can use after alignment. Here, we choose CHROM \cite{CHROM} to generate standard waves \footnote{Using more advanced methods can enhance the utilization, such as trained networks, while they may not be accessible for new trainers.}. Only those video slices with limited noise for methods used to generate standard labels are selected. From Fig \ref{Distribution-of-source2}, only half of datasets can be used for finally training process, not to mention only limited slices in qualified videos are highly convinced.

Like single heart rate, BYHE's label $R$ $\&$ output $\hat{R}$ either doesn't be affected by uncertain delay, and we needn't worry about the indefinite envelope shape of label waves. Meanwhile, $R$ transformed from ground truth waves contains more information than single heart rate and $R$ $\&$ $\hat{R}$ have clear physical meanings. As we analyzed in ``approach'', they share same value range and rhythm information. Hence when $\hat{R}$ is proved reasonable with $R$ through loss function (\ref{loss-func}), the end-to-end feature extractor can learn that they submit the correct temporal information as well. According to ablation study in Table \ref{Loss函数消融实验} and Fig \ref{Loss权重系数消融}, this connection between $R$ and $\hat{R}$ is proved reasonable. Therefore, we attribute performance improvement by BYHE to higher utilization on datasets and more effective label $\&$ output representation.

\section{Conclusion}
We investigate ubiquitous problems among rPPG datasets and propose an effective framework -- BYHE to solve it. According to analysis and experiments, we ascertain its reasonability and efficiency. By applying BYHE, we can conduct convenient end-to-end network training without necessity of struggling temporal alignment of label waves, and improve network's performance. One future direction is exploiting the usage of $\hat{R}$, such as combine them with non end-to-end methods (i.e. serve as input). Furthermore, there are still many obstacles waiting for us to conquer in front of efficient rPPG network training, such as (1) the unconscious usage of invalid data (videos contain very few rPPG information because of fierce noise, which can't be detected by human eyes) and (2) the lack of precise facial mask label which provides reliable position of rPPG signal (not all skin areas contain rPPG signals). In conclusion, although powerful feature extractor is important to end-to-end learning, more effective data processing and qualifying also need to be considered by researchers.

\appendix
\subsection*{Appendix.1. CWT filter $\&$ Preprocessing on ECG label wave}
CWT filter is the abbreviation of continuous wavelet transform filter, which is widely used in heart rate estimation field. Generally speaking, the fucntion of CWT filter is further refining the frequency on input waves, which can be viewed as heart rate. Next, we are going to exhibit how CWT filter works along with bandpass filter, some comrade researchers might be familiar with it.

\begin{figure}
    \centering
    \includegraphics[height=4.3cm]{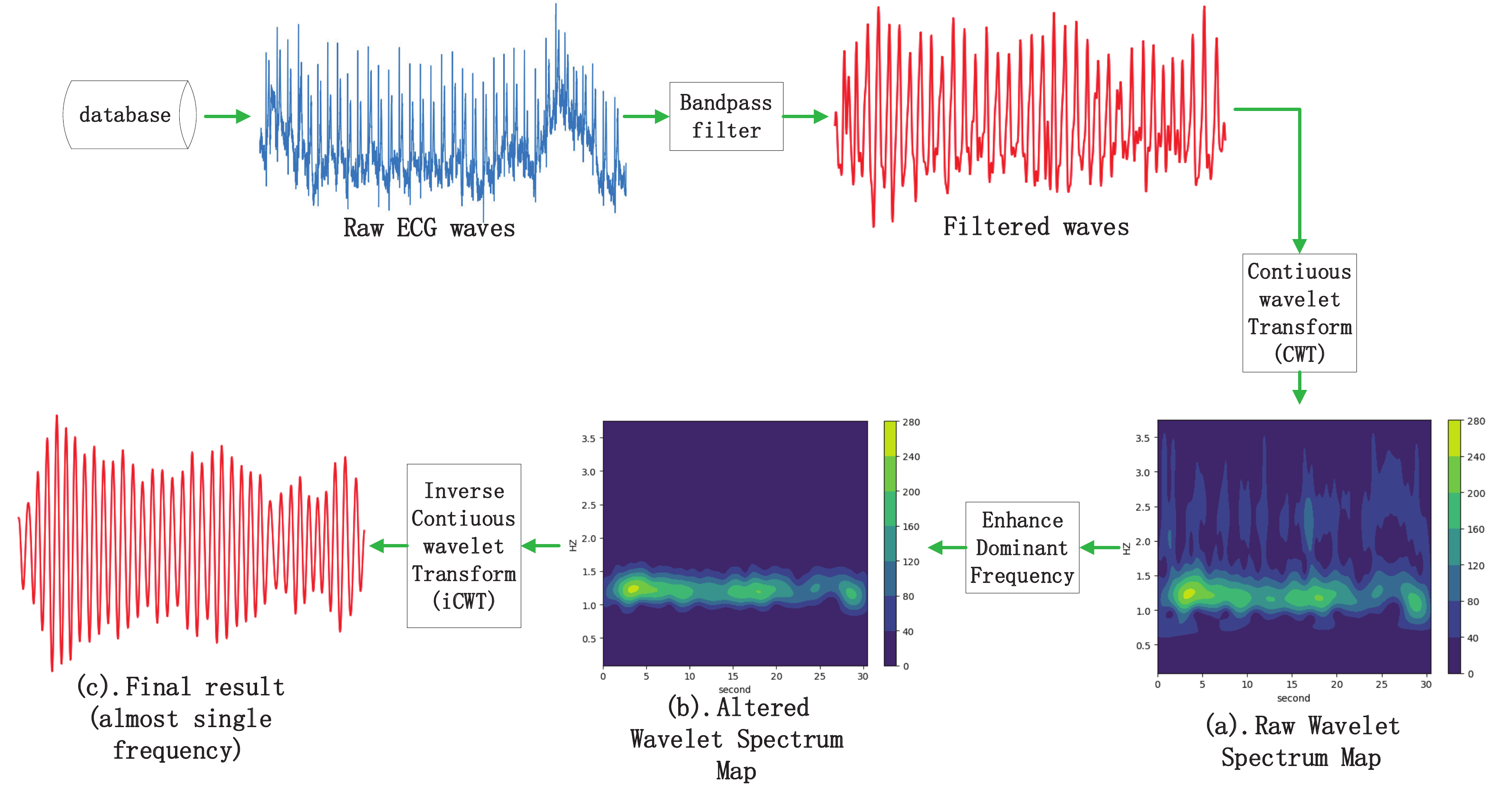}
    \caption{The procejure of bandpass filtering and CWT filtering on instance of ECG signals.}
    \label{cwt_filtering演示}
\end{figure}

Let's take a slice of electrocardiogram signal (ECG signals) as example here. Our additional experiments on HCI dataset (which adopts ECG signal as label) in main submission can prove that our method could be used on ECG signals. The specific process is shown in Fig.\ref{cwt_filtering演示}. Compared with Blood Volume Pulse (BVP) signal, ECG signal is relatively sticky when filtered because of QRS complex. Our involved CWT filter and bandpass filter can deal with the majority of ECG signals, but we still advise researchers to carefully check filtered waves when dealing with ECG signals.

First, we apply bandpass filter of [0.7Hz, 4Hz] on raw ECG signals (i.e $r(t)$ mentioned in main submission). The frequency of heart rate is always between 0.7Hz and 4Hz, hence using bandpass filter under such configuration could exclude some noise outside the frequency of heart rate. Then we apply CWT filter on filtered ECG signals. We use \textbf{Morlet wavelet basis} from 0Hz to 3.75Hz (which is also the frequency range of heart rate, generally speaking, the lower band and upper band of this filters is relatively flexible. All approximate lower band and upper band could be used, such as 0.3Hz to 3.5Hz, as long as the band could cover the main frequency of subject's heart rate.) And then we could generate a wavelet spectrum like (a) in Fig.\ref{cwt_filtering演示}. The horizontal axis stands for time (unit: ticks, depends on the samplingrate of electrocardiogram device) and the vertical axis stands for the corresponding ``frequency" of wavelet spectrum.

We calculate average value along ticks on each fixed ``frequency'' and pick the frequency corresponding to the largest average value, which is viewed as the dominant frequency of filtered label waves. Then we focus on it and gradually weaken the values of other frequencies by multiplying values on other frequencies by coefficient within 0 to 1. After iterating all ticks and doing operations above, we finally get the filtered wavelet spectrum, as shown in (b) in Fig.\ref{cwt_filtering演示}. It can be seen that main frequency is apparently more outstanding. Then we apply inverse CWT to turn the wavelet spectrum back to waves, which is the result of cwt filter (i.e. $f(t)$ mentioned in main submission). The result of CWT filter is shown as (c) in Fig.\ref{cwt_filtering演示}. Obviously, the filtered wave contains more monolithic frequency than input wave, and this frequency can be understood as heart rate. Filtered result of ECG signal resembles that of BVP siganls introduced (i.e. $f(t)$) in main submission, so the following preprocessing methods could be conducted successfully.

The CWT filter is quite important in our method: to extract stable phase change $A(t)$ using Hilbert transform, we hope the wave extracted contains as single frequency as possible (otherwise the phase change could be disturbed). Of course, we could use single heart rate to generate feature maps: directly generate a cosine wave whose frequency is the same with heart rate, and use it to perform Hilbert transform.

\subsection*{Appendix.2. Detailed Network Structure}
Detailed network structure is shown as Fig \ref{Detail_network} (can also be checked in Appendix.2). Compared with CAN in raw paper \cite{CAN}, we suitably increase the depth of CNN and shape of input facial frames, while we still keep its dual pipeline structure. All detailed convolution kernels are signed in Fig \ref{Detail_network}, and ``Skin Mask Fusion'' module refers to the configuration of raw paper (CAN $\&$ Deepphys) \cite{Deepphys, CAN}.

\begin{figure*}[t]
    \centering
    \setlength{\abovecaptionskip}{0.15cm}
    \includegraphics[width=18cm]{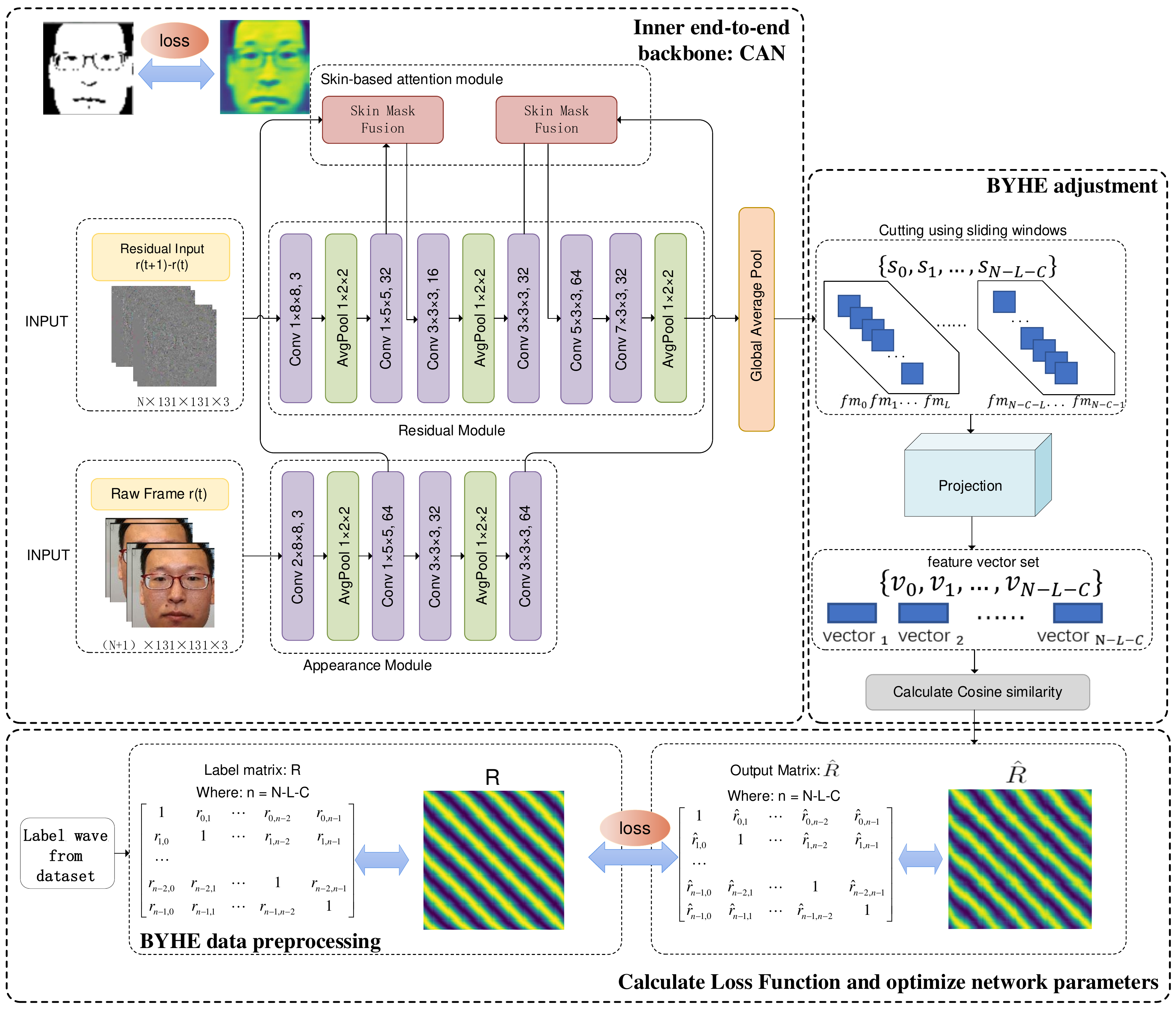}
    \caption{Detailed structure of network.}
    \label{Detail_network}
\end{figure*}

Pipeline dealing with residual inputs applies activition function ``tanh'' (We advise using activition function with symmetry property, which is conducive to final cosine similarity calculation.), and that dealing with raw inputs applies ``relu''. We don't set temporal padding in 3D CNN pipeline, as temporal padding does damage on generation of feature slices, adding the effect of sliding window, output length is shorter than input length. As shown in Fig \ref{Detail_network}, $fm_i$ represents the $i$th feature map along temporal dimension, $C$ represents the temporal convolution consumption (without padding) of inner end-to-end network, $L$ represents the temporal consumption of sliding window process (depends on the length of sliding window). This might be a little sticky, but in ``Experiment configuration'' mentioned below, we will exhibit how to compensate these consumed temporal frames.

\subsection*{Appendix.3. Experiment configuration}
Free from label wave alignment, the preparation for train label becomes convenient. When training, we select label waves at the same position with selected videos, as heart rate situation can be viewed constant under the assumption that the magnitude of uncertain delay are generally subtle comparing with heart rate variability. Then for selected label wave, we process it as proposed (Fig. 3 \textbf{in main submission}) and get generated label $\hat{R}$. As we mentioned above, because we don't set temporal padding and the mechanism of sliding window, we need to \textbf{increase input video's length} to compensate temporal consumption of BYHE, and what we do is covering additional frames both at the start and end of inputs (i.e. extra cover $(L+C)/2$ in beginning and ending).

Involved augmentation for input videos are commonly used in previous end-to-end method. Specifically, they are: (1) in 50$\%$ possibility perform horizontal and vertical flipping on input videos; (2) when heart rate of target instance lower than 70, in 50$\%$ possibility double sampling rate of input video slice and label wave using interpolation; (3) when heart rate of target instance higher than 90, in 50$\%$ possibility halve sampling rate of input video slice and label wave; (4) random slightly shake the face detection box (enlarge face detection box to 151x151x3, then random cut 131x131x3 slices on it).

We applied cosine dynamic learning rate \cite{COSINE_SGD} when conducting network training. The initial learning rate is 5e-4, min-learning rate is 1e-6, we don't use warm-up stragegy. Operating system is centOS 7, we use single GPU card ``Tesla v100'' with 16 GB memory which can just befit the batchsize of 6 (training process takes up about 15900 MB). In the beginning, we apply default parameter initialization method in pytorch to initialize network.

When performing ablation study to exhibit the performance of inner end-to-end network with out BYHE, we use CHROM \cite{CHROM} method to generate standard label waves, which following the configuration of pyVHR \cite{pyVHR} mentioned above. We apply CHROM on each video and CHROM returns generated predicted rPPG signal on that video. Then according to our observation, we pick those slices with high confidence (have obvious trait of heart rate information, which matches the frequency of corresponding label wave). Because CHROM generates rPPG signal directly based on the video, so its product can be viewed as temporally aligned with real rPPG signal. Then these high convinced generated rPPG signals are used as label wave to train inner end-to-end network. Of course, according to Fig. 8 \textbf{in main submission}, some videos are discarded after temporal wave alignment, especially those of v2, v3 and v9. Readers can refer to the manual of VIPL-HR datasets \cite{VIPL-HR}, where we can find that under these tasks, perticipants are required to make motion to produce noise. Such noise makes CHROM difficult to generate qualified and high convinced standard waves, thus results in the low utilization on these tasks.

\bibliography{aaai23}
\end{document}